\documentclass[conference]{IEEEtran}
\IEEEoverridecommandlockouts

\usepackage{amsfonts,amsmath,amsthm}
\usepackage{bm}
\usepackage{graphicx}
\usepackage{multirow}
\usepackage{subfig}
\usepackage{xcolor}
\usepackage{xspace}
\usepackage{cite}
\usepackage{amsmath,amssymb,amsfonts}
\usepackage{algorithmic}
\usepackage{graphicx}
\usepackage{textcomp}
\usepackage{xcolor}
\def\BibTeX{{\rm B\kern-.05em{\sc i\kern-.025em b}\kern-.08em
    T\kern-.1667em\lower.7ex\hbox{E}\kern-.125emX}}

\newcommand{\etal}{\emph{et~al.}\xspace}
\newcommand{\eg}{\emph{e.g.},\xspace}
\newcommand{\ie}{\emph{i.e.},\xspace}
\newcommand{\etc}{etc.\xspace}

\newcommand\figref[1]{Fig.~\ref{#1}}

\newcommand\tabref[1]{Table~\ref{#1}}

\newcommand\secref[1]{Sec.~\ref{#1}}
\newcommand\equref[1]{Eq.~(\ref{#1})}

\newcommand{\fakeparagraph}[1]{\vspace{1mm}\noindent\textbf{#1.}}

\newcommand{\sysname}{POSGen\xspace}

\begin{document}

\title{POSGen: Personalized Opening Sentence Generation for Online Insurance Sales}

\author{\IEEEauthorblockN{Yu Li\IEEEauthorrefmark{1},
Yi Zhang \IEEEauthorrefmark{2},
Weijia Wu \IEEEauthorrefmark{2},
Zimu Zhou   \IEEEauthorrefmark{3} and
Qiang Li \IEEEauthorrefmark{1}}
\IEEEauthorblockA{\IEEEauthorrefmark{1}College of Computer Science and Technology, Jilin University $\quad$ \IEEEauthorrefmark{2}Wesure Inc.}
\IEEEauthorblockA{\IEEEauthorrefmark{3}School of Information Systems, Singapore Management University}
Email: \{yuli19,li\_qiang\}@jlu.edu.cn,
\{jamesyzhang, wymanwu\}@wesure.cn,
zimuzhou@smu.edu.sg
}
\maketitle

\begin{abstract}
The insurance industry is shifting their sales mode from offline to online, in expectation to reach massive potential customers in the digitization era.
Due to the complexity and the nature of insurance products, a cost-effective online sales solution is to exploit chatbot AI to raise customers' attention and pass those with interests to human agents for further sales.
For high response and conversion rates of customers, it is crucial for the chatbot to initiate a conversation with personalized opening sentences, which are generated with user-specific topic selection and ordering.
Such personalized opening sentence generation is challenging because \textit{(i)} there are limited historical samples for conversation topic recommendation in online insurance sales and
\textit{(ii)} existing text generation schemes often fail to support customized topic ordering based on user preferences.
We design POSGen, a personalized opening sentence generation scheme dedicated for online insurance sales.
It transfers user embeddings learned from auxiliary online user behaviours to enhance conversation topic recommendation, and exploits a context management unit to arrange the recommended topics in user-specific ordering for opening sentence generation.
We conducted extensive offline tests to evaluate the performance of POSGen and also deployed it on a large online insurance platform.
POSGen outperforms the state-of-the-arts in offline tests and achieves 2.33x total insurance premium improvement through a two-month global test.
\end{abstract}


\section{Introduction}
\label{sec:intro}
The insurance industry is increasingly utilizing digital platforms for cost-effective product marketing and sales \cite{research2021global}.
For example, some insurance companies have deployed mini-programs on social media to approach large numbers of potential customers \cite{insuretech2019}.
To promote insurance sales via such digital channels, a chatbot can be exploited to initiate conversations with millions of users, while those interested are then seamlessly handed over to insurance agents for complex queries.



Although chatbot AI techniques have been widely applied for e-commerce services such as question answering \cite{braun2017evaluating}, chit-chat \cite{yan2018coupled}, product recommendation \cite{lei2020conversational} \etc, they are unfit for promoting online insurance sales due to the distinctive chatbot-user interactions therein.
Compared with other daily goods, insurance products are more complex and less familiar to the general public.
Many customers are reluctant to inquire on insurance products \cite{lim2009role}, and a tedious FAQ even scare potential buyers away \cite{graham20ux}.
It is believed that for insurance products, the chatbot should actively initiate the conversation with users via one or multiple \textit{opening sentences}, rather than passively waiting for their queries.
Effective opening sentences are crucial to raise user interests, which may eventually convert into sales \cite{de2019getting}.

\begin{figure}[t]
  \centering
  \includegraphics[width=0.42\textwidth, height=1.7in]{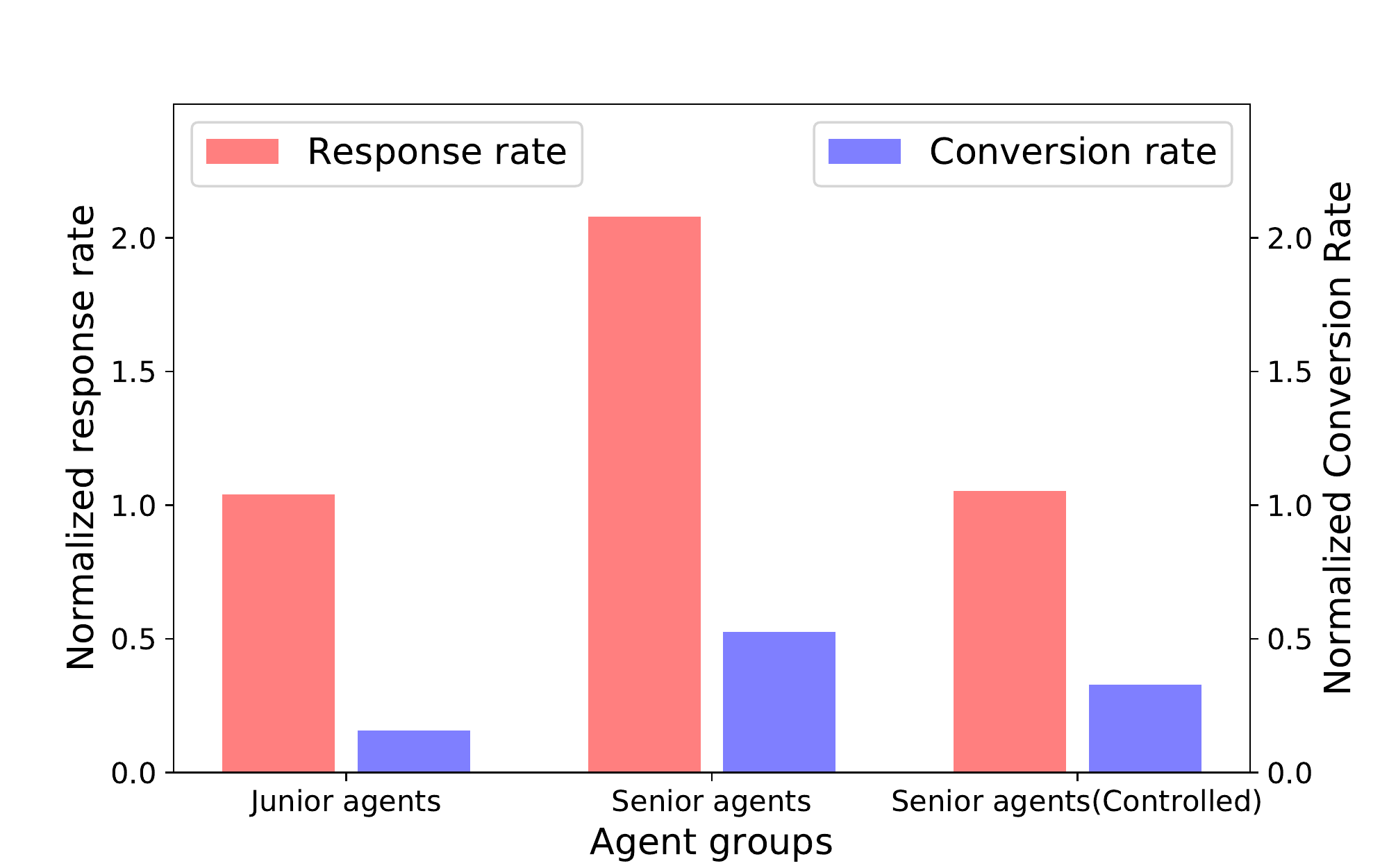}
  \caption{
  Effectiveness of opening sentences used by junior (within one year of experience), senior (over three years of experience) human agents and controlled senior agents (opening sentences are selected from junior agents).
  The customers are randomly distributed to the three agent groups.
  From the results we can see:
  \emph{(i)} Opening sentences from the senior group outperform the junior group in both response rate and conversion rate (junior agents vs senior agents).
  \emph{(ii)} Opening sentences is important in promoting final conversions (senior agents vs senior agents (controlled)).
  }
  \label{fig:group_comparison}
  \vspace{-0.2cm}
\end{figure}

\begin{figure}[t]
  \centering
  \includegraphics[width=0.42\textwidth]{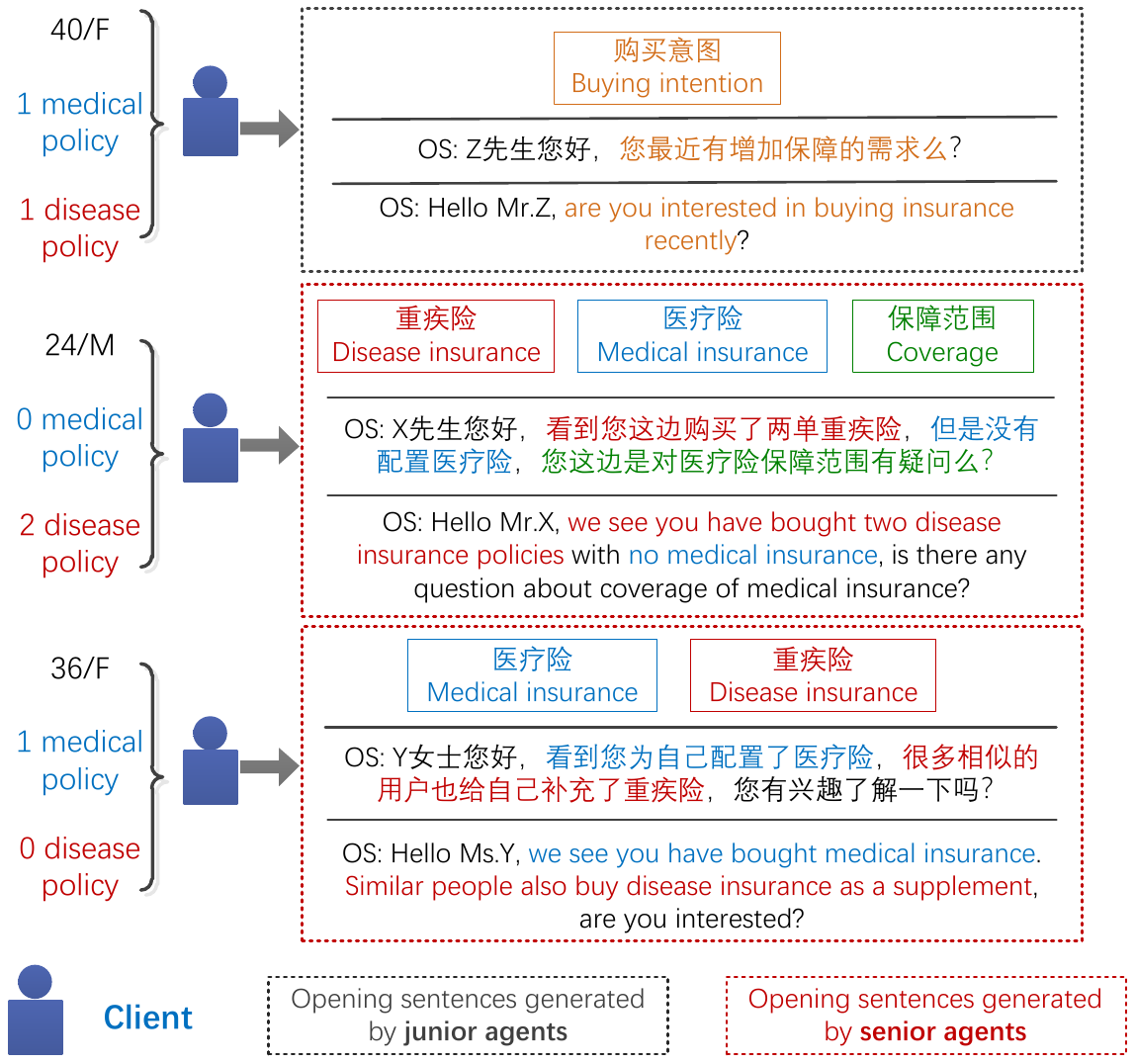}
  \caption{Example opening sentences used in the junior and senior agents. The left shows a few client features, which agents may exploit in his/her opening sentences.
  On the right shows the opening sentences and the corresponding topics.
  The senior agents tend to adjust the opening sentences according to the clients' features.}
  \label{fig:sent_comparison}
  \vspace{-0.2cm}
\end{figure}

To understand the characteristics of effective opening sentences, we resort to the historical opening sentences used by human agents and their sales records from a large online insurance platform.
We roughly divide the agents into two groups based on their seniority, \ie junior and senior.
In addition, we also introduce a controlled group of randomly selected senior agents for online tests, \ie senior agents (controlled), whose opening sentences are selected from junior agents by the system.
\figref{fig:group_comparison} plots the response rates (RRs, \ie the number of responded users divided by the number of contacted users) and the conversion rates (CVRs, \ie the number of converted users divided by the number of contacted users) of the opening sentences used by three agent groups.
The senior group notably outperforms the junior group in both RR and CVR, \ie the opening sentences used by the senior group are more effective.
Furthermore, the performances difference between senior agents and senior agents (controlled) demonstrate the opening sentences have significant impacts on conversions, even if agents have similar sales skills.
\figref{fig:sent_comparison} shows a few example opening sentences.
We observe that the opening sentences from the junior agents are generic and indifferent to customers.
In contrast, the effective opening sentences used by the senior group are highly \textit{personalized}: they exhibit user-specific topic selection and ordering.

However, personalized opening sentence generation for online insurance sales faces two technical challenges.
\begin{itemize}
    \item
    \textit{How to recommend user-specific topics for opening sentence generation?}
    Although it is viable to learn individuals' preferences on topics from the opening sentences used by senior agents and select the top $k$ for sentence generation, there are insufficient such historical conversation data.
    Given the vast user attributes and conversation topics, it is challenging to extract effective user representations for ranking the preference of topics given limited historical opening sentence samples.
    \item
    \textit{How to generate opening sentences with user-specific topic ordering?}
    This question belongs to the research on data-to-text generation \cite{dathathri2019plug, shu2020controllable, shen2019towards, yang2017generating, puduppully2019data}.
    However, prior studies either require fixed sentence length \cite{yang2017generating, li2018generating} or can only generate sentence from one topic or style \cite{shu2020controllable, dathathri2019plug}.
    A few \cite{shao2019long, shen2019towards,puduppully2019data} generate sentences on multiple topics, yet they arrange topics in a generic order \eg for fluency, without considering user-specific features.
\end{itemize}

To this end, we design \sysname, a personalized opening sentence generation scheme for promoting online insurance sales.
To address the data sparsity problem in topic recommendation, we exploit users' auxiliary behaviors (\eg clicks, conversions, and browsing history) about page items in the online insurance mini-program to enhance topic recommendation.
That is, these auxiliary behaviors on the same online insurance platform share similar user representations and are thus transferable to topic recommendation \cite{li2021revman, chen2019efficient}.
Accordingly, we design a selective attentive transfer learning model (SATL), which adaptively learns user representations from these auxiliary behaviors.
To generate sentences with user-specific topic ordering, we propose a context-aware sentence generation model (CSG).
In addition to the topics themselves, we also feed the user's embedding into a context manage unit to determine the next topic for sentence generation.
Thus the generated sentences not only account for fluency, but also conform to individual preferences.
Extensive real-world experiments demonstrate POSGen can significantly increase the insurance sales while also guarantee the quality of users' experiences.


Our contributions and results are summarized as follows:
\begin{itemize}
  \item
  To the best of our knowledge, \sysname is the first proposal on opening sentence generation dedicated to promote online insurance sales.
  \item
  We design a personalized opening sentence generation scheme with user-specific topic selection and ordering.
  We also tackle practical challenges such as data sparsity by learning transferable user representations from auxiliary user behaviors related to online insurance.
  \item
  We evaluate the performance of \sysname through both offline evaluation and online experiments.
  In the offline evaluation, we conducted comprehensive ablation studies with the state-of-the-arts to validate its effectiveness.
  Then we deployed \sysname on a large online insurance platform.
  Through a two-month global test, the results show that \sysname increased 51\% number of average talk rounds, which were finally converted into $2.33\times$ improvement on the total insurance premium without decreasing the quality of users' experiences (QoE).
\end{itemize}

In the rest of this paper, we review related work in \secref{sec:related}, present an overview of our solution in \secref{sec:design} and elaborate on its detailed designs in \secref{sec:SATL} and \secref{sec:CSG}.
We report both the offline and online evaluations in \secref{sec:offline} and \secref{sec:online}, and finally conclude in \secref{sec:conclusion}.

\section{Related Work}
\label{sec:related}
Our work is related to the following categories of research.

\subsection{Transfer Learning in Recommendation Systems}
Transfer learning transfers knowledge from the source to the target to improve the performance on source task, which often has limited data or supervision \cite{tan2018survey}.
It has been applied in recommendation systems since data sparsity is common in recommendation systems.
Zhao \etal \cite{zhao2013active} propose an entity-correspondence mapping method for collaborative filtering to improve recommendation quality.
Chen \etal \cite{chen2019efficient} devise an adaptive transfer network, with several attention gates to combine low-level feature representations from source domains to improve recommendation in the target domain.
However, these studies do not apply to our scenario.
They either require strong constraint on data in source and target domains, \eg the users are required to have features in source domain \cite{chen2019efficient}, or fail to design efficient structure for transferring knowledge from source to target \cite{li2020ddtcdr}.

The SATL model in our \sysname also exploits transfer learning to improve topic recommendation.
Unlike previous work, we relax the constraints on samples by designing a shared embedding layer to capture the co-occurrence of user features in the source and target domains.
Based on the shared embedding layer, we design a dual attentive mechanism in SATL to adaptively learn knowledge from source domains.

\subsection{Text Generation}
Personalized opening sentence generation falls into the area of data-to-text generation \cite{li2018generating, yang2017generating, sohn2015learning, shao2019long, dathathri2019plug, bowman2015generating}.
Bowman \etal \cite{bowman2015generating} propose a variation autoencoder (VAE) based method to generate sentences.
By introducing a latent variable, the model can generate sentences with diverse patterns.
A follow-up \cite{sohn2015learning} introduces a context prior to generate sentences with given contexts, \eg topics.
PHVM \cite{li2018generating} adapts the model in \cite{sohn2015learning} for generating product descriptions.
Other applications of text generation include controllable poetry generation \cite{yang2017generating}, data-to-summary generation \cite{puduppully2019data}, \etc

Our personalized opening sentence generation is more challenging because we demand user-specific sentence structuring, which no prior studies can fulfill.
Hence we propose a context-aware sentence generation model (CSG) to adjust the topic ordering based on user features.


\section{\sysname Overview}
\label{sec:design}
\begin{figure}[t]
  \centering
  \includegraphics[width=0.4\textwidth, height=3in]{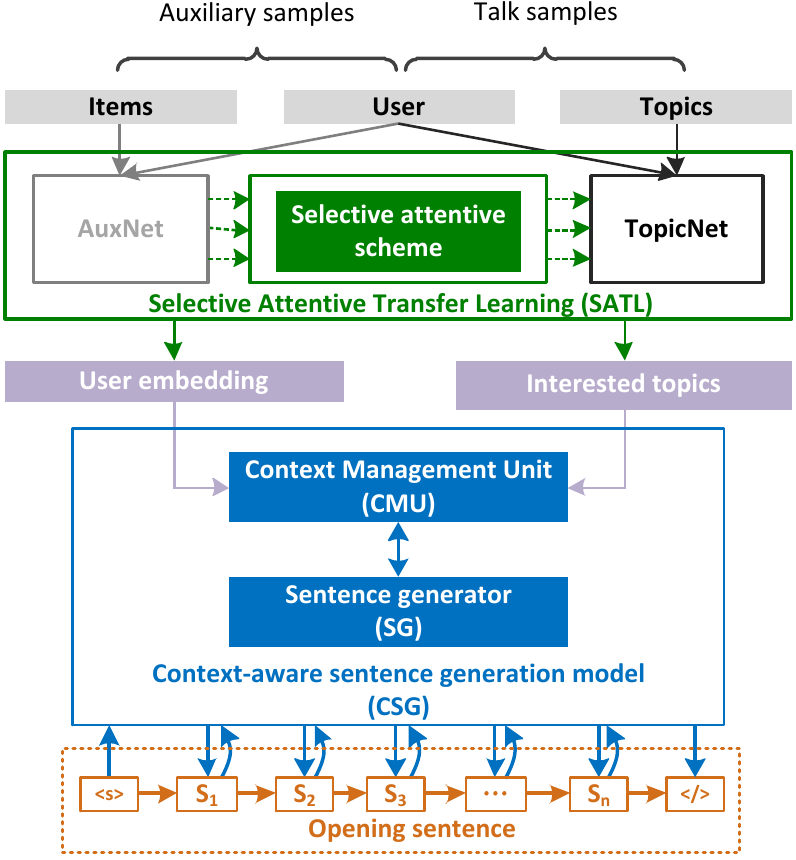}
  \caption{\sysname overview.
  It has two components: selective attention transfer learning (SATL) and context-aware sentence generation model (CSG).
  SATL accounts for extracting user's embedding and recommending interested topics.
  The results of SATL are then passed to CSG to generate personalized opening sentences.}
  \label{fig:workflow}
\end{figure}

In this section, we present an overview of \sysname, a new personalized opening sentence generation scheme for promoting online insurance sales.
As shown in \figref{fig:workflow}, \sysname consists of two functional modules.
\begin{itemize}
  \item \textbf{Selective attentive transfer learning model (SATL).}
  This module extracts user embeddings and recommend the top $k$ conversation topics via transferring learning from auxiliary user behaviours like clicks on presented insurance products on the same online insurance platform.
  SATL is a two-tier model consisting of an auxiliary network (AuxNet) for training auxiliary samples and a topic recommendation network (TopicNet) for topic recommendation.
  TopicNet shares feature embedding with AuxNet and uses a dual attention array to adaptively learn high-level knowledge from AuxNet.
  The design enables TopicNet to learn comprehensive user preferences for accurate topic recommendation.
  \item \textbf{Context-aware sentence generation model (CSG).}
  This module generates personalized sentences based on the given user embedding and recommended topics from SATL.
  In CSG, we design a context management unit to customize topic orders based on user's embedding.
  The ordered topics then guide the sentence generator and output personalized opening sentences.
\end{itemize}

\section{Selective Attentive Transfer Learning (SATL)}
\label{sec:SATL}

\begin{figure}[t]
  \centering
  \includegraphics[width=0.44\textwidth]{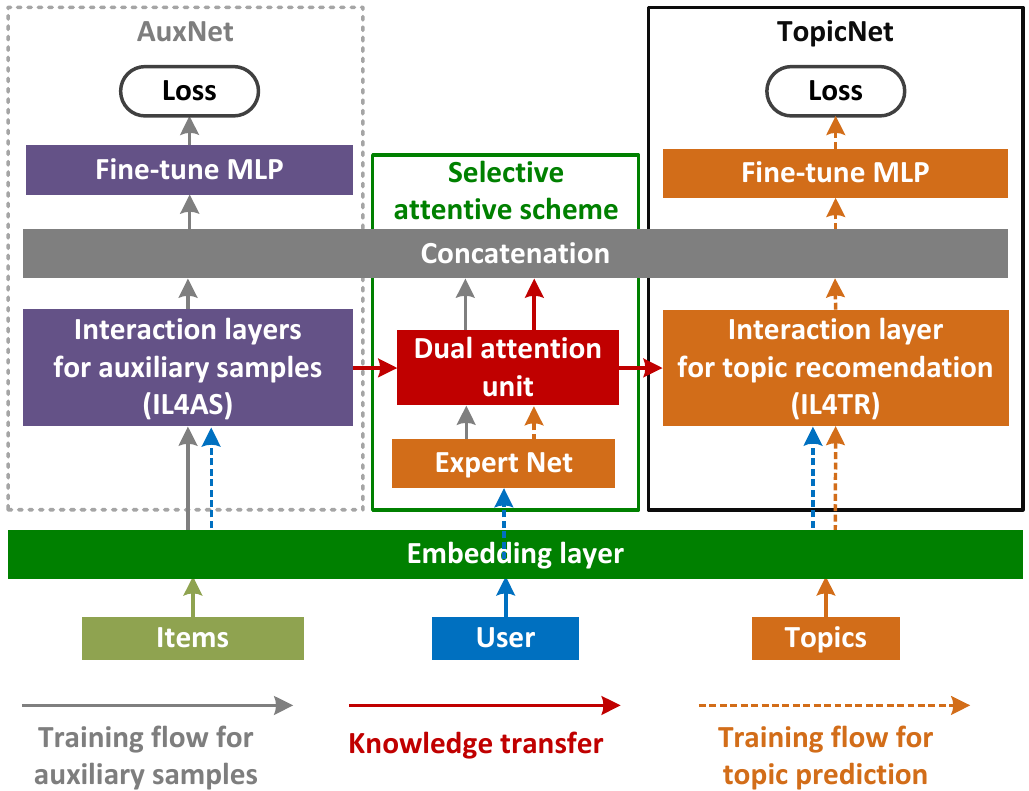}
  \caption{SATL architecture.
  A dual attention array cooperates with an expert network to transfer knowledge from auxiliary samples to topic recommendation.}
  \label{fig:SATL}
\end{figure}

This section presents the design of our selective attentive transfer learning (SATL) model, which aims to predict users' topics of interest.
Given the sparsity of talk samples, we design a selective attentive scheme to transfer knowledge from samples of users' auxiliary behaviours.
With the assist of the transferred knowledge, SATL is able to capture rich users' representations and make accurate topic prediction.
\figref{fig:SATL} shows the architecture of SATL.
We explain its data input and knowledge transfer mechanism below.


\subsection{Data Input}
SATL takes two types of data inputs: auxiliary samples and talk samples.
The talk samples are collected from the conversations between agents and customers.
The topics in the talk samples are manually annotated.
The auxiliary samples have two components: item features and user features.
Items refer to the elements users interact with, \eg presented page item when user clicks.
Talk samples consist of user features and topics collected from opening sentences.
The auxiliary samples are obtained from three user behaviours on the same insurance platform: clicks, conversions, and visit histories, with 0/1 as the labels.
\tabref{table:aux_behaviors} lists the items and descriptions of each behavior of the auxiliary samples.
For each type of samples, we mark its label as 1 if the user clicks/buys/visits the given item, and 0 otherwise.

\begin{table}[t]
\caption{Summary of auxiliary behaviors.}
\centering
\resizebox{\columnwidth}{!}{%
\begin{tabular}{lll}
\hline
\textbf{Behavior types} & \textbf{Items}  & \textbf{Description}           \\ \hline
Click              & insurance/videos/articles & Whether a user clicks an item \\ \hline
Conversion & insurance & \begin{tabular}[c]{@{}l@{}}Whether a user buys an\\  insurance product\end{tabular} \\ \hline
Visit history      & insurance/videos/articles & Whether a user visits an item   \\ \hline
\end{tabular}%
}
\label{table:aux_behaviors}
\end{table}

We use a multi-hot vector $\hat{q}_{ik}$ to denote the auxiliary sample of user $i$ with item $k$, \ie
\begin{equation}
\begin{aligned}
\hat{q}_{ik}=[\underbrace{1,0,...,1}_{user\ \hat{u}_i},\underbrace{0,1,...,0}_{item\ \hat{im}_k}]
\end{aligned}
\end{equation}
where $im_k$ refers to the k-th item.
The dense vector $q_{ik}$ embedded by the dictionary is as $q_{ik}=Emb(\hat{q}_{ik})$.
Similarly, we use $\hat{x}_{jl}=[\hat{u}_j, \hat{tp}_l]$ for the corresponding raw talk sample of user $j$ and topic $l$.


\subsection{Knowledge Transfer Mechanism}
SATL is inspired by knowledge transfer in recommendation systems \cite{ma2018entire}.
As with these proposals, the embedding layer of SATL is shared for both tasks since those users are from the same online insurance platform and share similar feature representations.
However, SATL improves upon \cite{ma2018entire} by allowing not only such low-level feature sharing, but also high-level feature sharing.
Specifically, we design a dual attention array and an expert network in the interaction layers between the AuxNet and TopicNet.
The expert network is activated by both auxiliary samples and talk samples for learning more representative user embeddings.
Upon the expert network, a dual attention array is designed for extracting high-level features to improve the accuracy of topic recommendation task.
For the expert network, we use multi-layer perceptron as its inner structure.
In the dual attention array, we design two attention units \ie $\bm{Att}=[Att^{aux},Att^{tp}]$.
$Att^{aux}$ controls the weight on the expert network output when trained on auxiliary samples, while $Att^{tp}$ controls the weight on the expert network output when trained on talk samples.

To explain the details of the high-level knowledge transfer, we abstract the interaction layers of the the AuxNet and TopicNet as interaction layers for auxiliary samples (IL4AS), and interaction layers for topic prediction (IL4TR), respectively.
We use $H$ to denote the intermediate output of IL4AS and $G$ for IL4TR.
In the training process, we first train IL4AS through auxiliary samples and then, train IL4TR through talk samples and trained IL4AS.
Let the output of expert network be $F(u_i)$.
The attention weights $\bm{w}^{aux}$ for auxiliary sample $q_{ik}$ is computed as:
\begin{equation}
\begin{aligned}
\bm{a}^{aux}&=H(q_{ik})*W^{aux}*F(u_i)^T \\
\bm{w}^{aux}&=softmax(\textbf{a}^{aux}) \\
\end{aligned}
\label{eq:att_weight}
\end{equation}
where $W$ is the weight matrix in the attention unit.
The output of $Att^{aux}$ is then $Att^{aux}(q_{ik}, F, H)=\bm{w}^{aux}F(u_i)$.

The training for topic recommendation is similar.
The only difference is that we use the output of the previously trained expert network through dual attention array.
Following the definition above, we have:
\begin{equation}
\begin{aligned}
Att^{tp}(x_{jl}, F, G)=\bm{w}^{tp}F(u_j)
\end{aligned}
\end{equation}

The output of dual attention array is concatenated with $H$ and $G$ respectively and then sent to the upper fine-tune MLP to calculate the loss.
We use two loss functions: $\mathcal{L}^{aux}$ for auxiliary samples and $\mathcal{L}^{tp}$ for topic recommendation.
For simplification, we use logloss for the two loss functions.
The loss of SATL $\mathcal{L}$ can be calculated as:
\begin{equation}
\begin{aligned}
\mathcal{L} &\!=\! \alpha\mathcal{L}^{aux} + (1-\alpha)\mathcal{L}^{tp}+\lambda\Omega(\Theta) \\
\end{aligned}
\end{equation}
where $\alpha$ is pre-defined controlling weights, $\Omega(\Theta)$ is the regularization for network parameters and $y$ is the sample label.
For $\mathcal{L}^{aux}$ and $\mathcal{L}^{tp}$, CE denotes the cross entropy and MLP denotes the fune-tune multi-layer perceptron.

\section{Context-aware Sentence Generation (CSG)}
\label{sec:CSG}

\begin{figure}[t]
  \centering
  \includegraphics[width=0.44\textwidth]{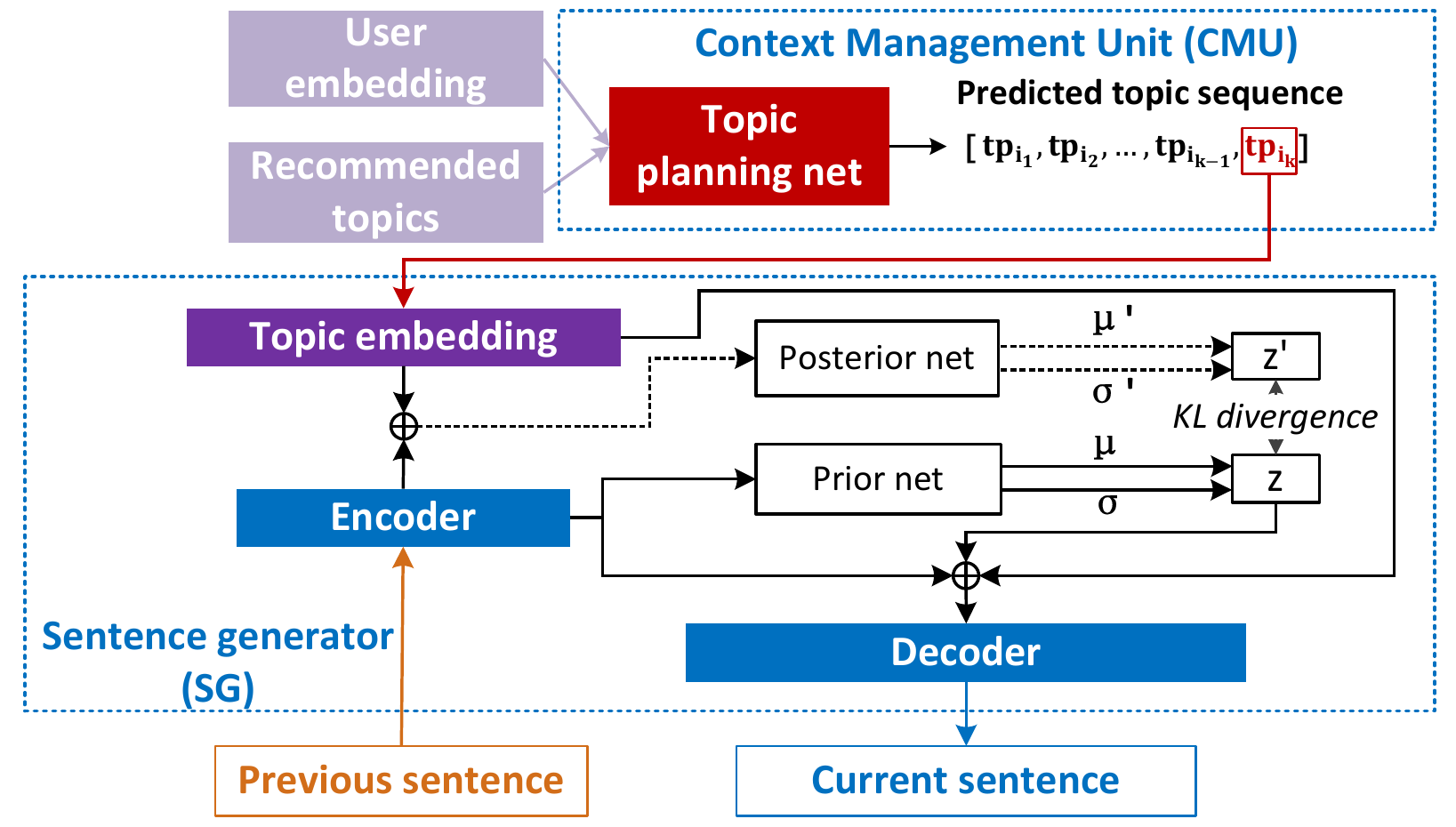}
  \caption{CSG architecture.
  A context management unit (CMU) is used to customize the topic order based on user's embedding.
  The recommended topics are then sent to sentence generator (SG) to generate opening sentences.}
  \label{fig:CSG}
\end{figure}

\begin{figure}[t]
  \centering
  \includegraphics[width=0.45\textwidth]{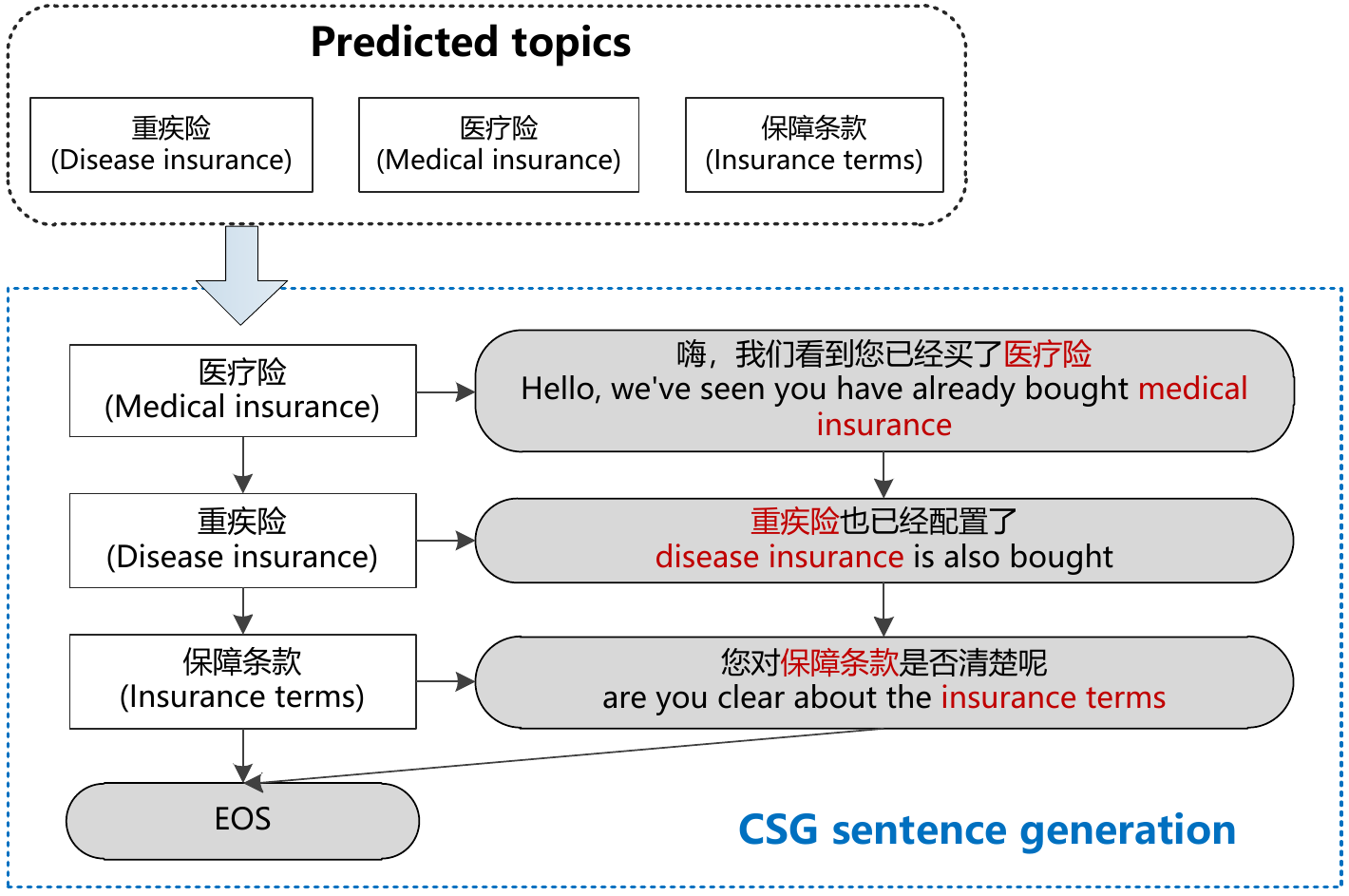}
  \caption{An example of generation process in CSG. CSG first plans the order of predicted topics.
  Then it conducts sentence generation conditioned on the target topic and previous generated sentences.}
  \label{fig:CSG_example}
\end{figure}

This section introduces the context-aware sentence generation (CSG) model, which focuses on generating opening sentences based on the predicted topics from SATL.
To achieve this goal, we design two components in CSG: a context management unit (CMU) and a sentence generator (SG).
The CMU determines the ordering of topics so that the generated sentences is logical and easy to understand.
The generated topic sequences are then passed to SG to generate corresponding sentences.
In SG, we design a conditional VAE \cite{sohn2015learning} for sentence generation.
Different from previous works, we incorporate previous sentence in the context so that the fluency and coherence of generated sentences can be guaranteed.
\figref{fig:CSG} shows the architecture of CSG.
For better illustration of sentence generation process, we also demonstrate a generation example in \figref{fig:CSG_example}.

\subsection{Context Management Unit (CMU)}
This module is designed for planning topic order based on user's preferences, \ie user's embedding from SATL.
The input of CMU is user's embedding and the recommended topics from SATL.
The input will be sent to a topic planning network to generate topic sequence based on user's preferences.
We design three components in the topic planning network: a GRU unit, an attention unit and a fine-tune MLP.
Let the topic sequence by step $n-1$ be $\bm{\text{TP}}_{n-1}=\{tp_{i_1},tp_{i_2},...,tp_{i_{n-1}}\}$.
The process in generating $\bm{\text{TP}}_{n}$ is as below.

First, we use the embedding dict, \ie the embedding layer from SATL, to derive the dense vector of $TP_{n-1}$, \ie $v^{TP}_{n-1}$.
The topic sequence vector $v^{TP}_{n-1}$ are then sent to GRU to get encoded output $O_n^{TP}$:
\begin{equation}
\begin{aligned}
O^{TP}_n, h_n^{TP}&=\bm{GRU}^{\text{CMU}}(v^{TP}_{n-1}, h^{TP}_{n-1})
\end{aligned}
\end{equation}
where $h^{TP}_{n-1}$ is hidden state for step $n-1$.
When $n=1$, we pad $h^{TP}_0$ with zeros for initialization.

We then feed the output $O^{TP}_{n}$ and user embedding $u$ into the attention unit to calculate the tuned vector, \ie $Att^{\text{CMU}}_{n}$
\begin{equation}
\begin{aligned}
Att^{\text{CMU}}_{n}=softmax(u*W^{\text{CMU}}*O^{TP}_{n})*O^{TP}_{n} \\
\end{aligned}
\end{equation}
Afterwards, the output $Att^{\text{CMU}}_{n}$ is concatenated with $u$ and sent into the MLP to calculate the probability on recommended topics, \ie
\begin{equation}
\begin{aligned}
p(tp_i|u, \bm{TP}_{n-1}) = \bm{MLP}^{\text{CMU}}([Att^{\text{CMU}}_{n}; u])
\end{aligned}
\end{equation}
Following \cite{shen2019towards,shao2019long}, we use beam search to generate the topic sequence.

\subsection{Sentence Generator (SG)}
This module takes the recommended topic sequence $\bm{\text{TP}}_n$ as input and generates personalized opening sentences.
We build SG on top of a conditional variational autoencoder (CVAE) for its capability to generate controlled sentences \cite{sohn2015learning}.
Let $\bm{S}_{n-1}=\{w_1,w_2,...,w_M\}$ be the sentence generated based on $\bm{\text{TP}}_{n-1}$.
The process to generate $\bm{S}_{n}$ is as follows.

To generate $\bm{S}_{n}$ based on the topic sequence as well as guarantee its quality, \eg coherence, we use topic $tp_{i_n}$ and previous sentence $\bm{S}_{n-1}$ as the context $c$ to learn the latent factor for generation, \ie
\begin{equation}
\begin{aligned}
c=[Emb(tp_{i_n}); enc(\bm{S}_{n-1})]
\end{aligned}
\end{equation}
where $enc(\bm{S}_{n-1})$ denotes the embedding vector from the encoder.
As in \cite{sohn2015learning}, we design two networks, \ie posterior network and prior network, to approximate the posterior $q_{\theta}(z|x,c)$ and prior $p_{\phi}(z|c)$
\begin{equation}
\begin{aligned}
&Net_{\text{Post}}(enc(\bm{S}_{n-1}), c)\!=\![\mu'; \sigma'] \\
&Net_{\text{Prior}}([Emb(tp_{i_n});enc(\bm{S}_{n-1})])\!=\![\mu; \sigma] \\
\end{aligned}
\end{equation}
where $enc(\bm{S}_{n-1})$ is the encoded vector of $\bm{S}_{n-1}$.
Then the latent parameter $z'$ for posterior network and $z$ for prior network can be calculated via reparameterization, \ie $z'=\mu'+\epsilon'*\sigma', \  \epsilon'\in\mathcal{N}(0,1)$, $z=\mu+\epsilon*\sigma, \  \epsilon\in\mathcal{N} (0,1)$.
The design enables posterior net to learn a more accurate distribution of $z$ to generate $\bm{S}_n$
given the connection between $\bm{S}_n$ and $\bm{S}_{n-1}$.
In addition, the concatenation makes it feasible for the latent factor $z$ to capture the smooth transition between $\bm{S}_{n-1}$ and $\bm{S}_n$ from training.

The latent variable $z$ is then sent to the decoder for sentence generation.
In our design, we choose bidirectional GRU as the inner structure of the decoder.
We concatenate the topic embedding $Emb(tp_{n})$, encoded previous sentence vector $enc(\bm{S}_{n-1})$ and the latent vector $z$ as the initial hidden state of the decoder.
Denoting $dec$ as the output of decoder, we then greedily generate words, \ie
\begin{equation}
\begin{aligned}
w^{\bm{S}_n}_i&=\mathop{\arg\max}_{i}P(w^{\bm{S}_n}_i|dec(\{w^{\bm{S}_n}_{<i}\}_{(k)}, h^{dec}_n)) \\
h^{dec}_n&=
[emb(tp_{i_n});\ enc(\bm{S}_{n-1});\ z_{n}]
\end{aligned}
\end{equation}
where $\{w^{\bm{S}_n}_{<i}\}$ denotes the previous word sequence before step $i$.
When $n=0$, we pad $enc(\bm{S}_{n-1})$ as zeros in $h^{dec}_n$.


\subsection{Loss Function of CSG}
Based on the structure of CSG, the topic sequence generation guides the sentence generation.
Accordingly, we can split the training process of CSG into two phases and use two loss functions for CMU and SG.

We use cumulated softmax loss as the loss function for CMU.
The loss function $\mathcal{L}^{\text{CMU}}$ for topic sequence $\bm{TP}_i$ is
\begin{equation}
\begin{aligned}
\mathcal{L}^{\text{CMU}}(\bm{TP}_i)=-\frac{1}{K}\sum_{k=1}^K\sum_{n=1}^{N^{tp}}y_{i,k}(m)\log(tp_{i,k}(m))
\end{aligned}
\end{equation}
where K is the topic length of this sample, $N^{tp}$ is the number of topic candidates, and $y_{i,k}$ is the $k$-th topic label, \ie $y_{i,k}(m)=1$ if the $m$-th topic in $N^{tp}$ is at the $k$-th place.

For SG, we follow the design in \cite{sohn2015learning} and use empirical lower bound (ELBO) as the loss:
\begin{equation}
\begin{aligned}
\mathcal{L}^{\text{SG}}(\bm{S})&=\beta\mathcal{L}_{KL}+\mathcal{L}_{pred}\\
\end{aligned}
\end{equation}
where $\mathcal{L}^{\text{SG}}$ contains two parts: the loss from KL divergence $\mathcal{L}_{KL}$ and, the loss of reconstructing $\bm{S}_i$ based on the context c, \ie $\mathcal{L}_{pred}$:
\begin{equation}
\begin{aligned}
\mathcal{L}_{KL}&=-KL(q_{\theta}(\bm{z}|\bm{enc(S)},\bm{c})||p_{\phi}(\bm{z}|\bm{c})) \\
\mathcal{L}_{pred}&=\frac{1}{|\bm{S}|}\sum_{i=1}^{|\bm{S}|}\sum_{j=1}^{|\bm{S}_{i}|}\log p_{\theta}(w^{\bm{S}_{i}}_j|w^{\bm{S}_{i}}_{<j},\bm{S}_{i-1},z_i,tp_i)
\end{aligned}
\end{equation}
We design $\beta$ as a dynamic weight on $\mathcal{L}_{KL}$ for mitigating KL vanishing problem \cite{sohn2015learning}.
The detailed controlling scheme will be shown in the evaluation.


\section{Offline Evaluations}
\label{sec:offline}

This section reports the offline evaluations of \sysname.
Specifically, we assess \sysname's performance on topic recommendation and sentence generation separately by comparing the corresponding functional modules \ie SATL and CSG, with the state-of-the-arts.


\subsection{Experimental Setups}

\fakeparagraph{Datasets}
We use two datasets for offline evaluation.
One is the ads description from \cite{shao2019long} and the other is the insurance data collected from an online insurance platform.
The ads dataset has 119k samples while the insurance dataset contains 2000k user behavior samples and 80k talk samples.
The sensitive information of all samples, \eg user name and phone number, is concealed for privacy preservation.
The insurance dataset is for topic recommendation.
The ads and talk samples are for sentence generation.

\begin{table}[t]
\caption{Parameters used in SATL.}
\centering
\resizebox{\columnwidth}{!}{%
\begin{tabular}{|l|l|l|}
\hline
\multicolumn{1}{|c|}{\textbf{Parameter Name}} & \multicolumn{1}{c|}{\textbf{Description}}                                                                       & \multicolumn{1}{c|}{\textbf{Value}} \\ \hline
private\_dnn\_layer                           & \begin{tabular}[c]{@{}l@{}}number of dnn layers for each \\ \ie AuxNet and TopicNet\end{tabular} & 6                                   \\ \hline
expert\_dnn\_layer                            & number of layers for expert net                                                                                 & 4                                   \\ \hline
final\_dnn\_layer                             & layers of fine-tune MLP                                                                                         & 2                                   \\ \hline
expert\_num                                   & number of subnets in expert networks                                                                            & 16                                  \\ \hline
emb\_size                                     & embedding size                                                                                                  & 128                                 \\ \hline
check\_period                                 & validation period                                                                                               & 50                                  \\ \hline
\end{tabular}%
}
\label{table:satl_parameter}
\end{table}

\fakeparagraph{Model Parameter Configurations}
For reproducibility, we summarize the important model parameters in each component of \sysname.
\tabref{table:satl_parameter} lists the parameters of SATL.
In SATL, we use symmetric parameter settings for AuxNet and Topic to improve the efficiency of tuning.
For CSG, we use two tables to show the parameters of its two components, \ie CMU and SG.
The parameters of context management unit (CMU) are shown in \tabref{table:cmu_parameter}.
It can be treated as a simpler version of SATL and some of its parameters are copied from SATL.
The different parameters are \emph{stop\_prob} and \emph{max\_topic\_len}.
Those two parameters are used for determining the stopping condition of topic sentence generation, \ie when the max predicted probability of next topic is less than \emph{stop\_prob} or the length of generated topic sequence reaches \emph{max\_topic\_len}.
In practice, we set \emph{max\_topic\_len}=5.
The parameters of sentence generator (SG) are listed in \tabref{table:sg_parameter}.
As listed, there are two parts of parameters: one for the CVAE model and another for dynamic control on $\beta$, \ie
\begin{equation}
\begin{aligned}
\beta&=\frac{1}{1+\exp(-k*((i-sp_0)\ \textbf{mod}\ cycle))}
\end{aligned}
\label{eq:beta_control}
\end{equation}
The control scheme is for mitigating KL-vanishing problem.
Details will be shown in \secref{sec:study_sg}.

\begin{table}[t]
\caption{Parameters used in CMU.}
\centering
\resizebox{\columnwidth}{!}{%
\begin{tabular}{|l|l|l|}
\hline
\multicolumn{1}{|c|}{\textbf{Parameter Name}} & \multicolumn{1}{c|}{\textbf{Description}} & \multicolumn{1}{c|}{\textbf{Value}} \\ \hline
user\_emb\_size                               & embedding size of users                   & 128                                 \\ \hline
item\_emb\_size                               & emb size of topic                         & 128                                  \\ \hline
stop\_prob                                   & stop probability for topic generation              & 4e-2                                 \\ \hline
max\_topic\_len                                   & max length of topic sequence              & 5                                 \\ \hline
\end{tabular}%
}
\label{table:cmu_parameter}
\end{table}

\begin{table}[t]
\caption{Parameters used in CSG.}
\centering
\resizebox{\columnwidth}{!}{%
\begin{tabular}{|l|l|l|}
\hline
\multicolumn{1}{|c|}{\textbf{Parameter Name}} & \multicolumn{1}{c|}{\textbf{Description}}                                                               & \multicolumn{1}{c|}{\textbf{Value}} \\ \hline
max\_doc\_len                                 & max length of a sentence                                                                                & 100                                 \\ \hline
emb\_size                                     & emb size of sentence                                                                                    & 100                                 \\ \hline
latent\_size                                  & size of latent factor for z                                                                                   & 300                                 \\ \hline
dropout\_rate                                 & dropout\_rate                                                                                           & 0.1                                 \\ \hline
hidden\_size                                  & size of hidden layer of encoder and decoder                                                                                  & 64                                  \\ \hline
num\_layers                                   & number of layers                                                                                        & 4                                   \\ \hline
cycle                                         & \multirow{3}{*}{\begin{tabular}[c]{@{}l@{}}annealing control parameters \\ for KL-LOSS\end{tabular}}    & 4000                                \\ \cline{1-1} \cline{3-3}
k                                             &                                                                                                         & 0.003                                \\ \cline{1-1} \cline{3-3}
$sp_0$                                    &                                                                                                         & 1000                                \\ \hline
\end{tabular}%
}
\label{table:sg_parameter}
\end{table}



\subsection{Ablation Studies on Topic Recommendation}
In this subsection, we compare the performance of SATL with the state-of-the-arts on topic recommendation in standard benchmarks.

\fakeparagraph{Baselines}
We compare SATL with the following baselines.
\begin{itemize}
  \item Logistic regression \cite{mcmahan2013ad}.
  This method serves as the baseline.
  To show the benefit of transferred knowledge, LR is only trained on the talk samples.
  \item CoNet \cite{Hu2018CoNet}.
  It utilizes a sparse matrix for transferring knowledge from source domain to target domain.
  Following its design, matrix is designed for both interaction layer and fine-tune MLP.
  \item DDTCTR \cite{pan2017transfer}.
  It proposes an orthogonal transformation mechanism to transfer knowledge.
  We modify its loss function to logloss in our scenario.
\end{itemize}

\fakeparagraph{Metrics}
We use the standard metrics in topic recommendation.
\begin{itemize}
    \item AUC:
    This metric stands for ``area under the curve''.
    It is a metric estimating the performance of a classification model at all classification thresholds.
    \item Recall@N:
    This metric is based on primary evaluation ``Recall''.
    Recall@N means we check the rate if positive items appear in top N candidates.
    \item NDCG:
    This metric stands for normalized discounted cumulative gain and serves as a measure on the quality of recommended lists.
    It assigns higher scores to the positives items if they get higher ranks in the list.
\end{itemize}

\fakeparagraph{Setups}
We randomly split the data into two groups: 80\% for training \ie around 1600k auxiliary samples and 64k talk samples, and the other 20\% for testing.
By default, we select the top 10 topics.

\fakeparagraph{Results}
\tabref{table:topic_rec_performance} summarizes the performance of different methods on topic recommendation.
Compared with the best baselines, SATL is the best in all metrics, \ie the relative improvements are 3.89\% in AUC, 10.8\% in NDCG and 4.5\% in Recall.
The results show that both the high-level knowledge and the low-level feature representation can be efficiently transferred from auxiliary samples.
Thus SATL learns richer user preferences and achieves more accurate topic recommendation.

\begin{table}[t]
\centering
\caption{Performance of topic recommendation. Relative improvement with the best candidate is shown in the last row.}
\begin{tabular}{l|l|l|l}
\hline
\textbf{Model}                                          & \textbf{AUC}                                             & \textbf{NDCG@10}                                             & \textbf{Recall@10}                                         \\ \hline
LR                                                      & 0.685                                                    & 0.315                                                     & 0.692                                                       \\ \hline
CoNet                                                   & 0.75                                                     & 0.3932                                                    & 0.7751                                                     \\ \hline
DDTCTR                                                  & 0.77                                                     & 0.4541                                                    & 0.8381                                                     \\ \hline
\begin{tabular}[c]{@{}l@{}}SATL\\ (RelImp)\end{tabular} & \begin{tabular}[c]{@{}l@{}}0.80\\ (\textbf{+3.89\%})\end{tabular} & \begin{tabular}[c]{@{}l@{}}0.5032\\ (\textbf{+10.8\%})\end{tabular} & \begin{tabular}[c]{@{}l@{}}0.8761\\ (\textbf{+4.5\%})\end{tabular} \\ \hline
\end{tabular}
\label{table:topic_rec_performance}
\end{table}

\begin{figure}[t]
  \centering
  \includegraphics[width=0.43\textwidth, height = 1.7 in]{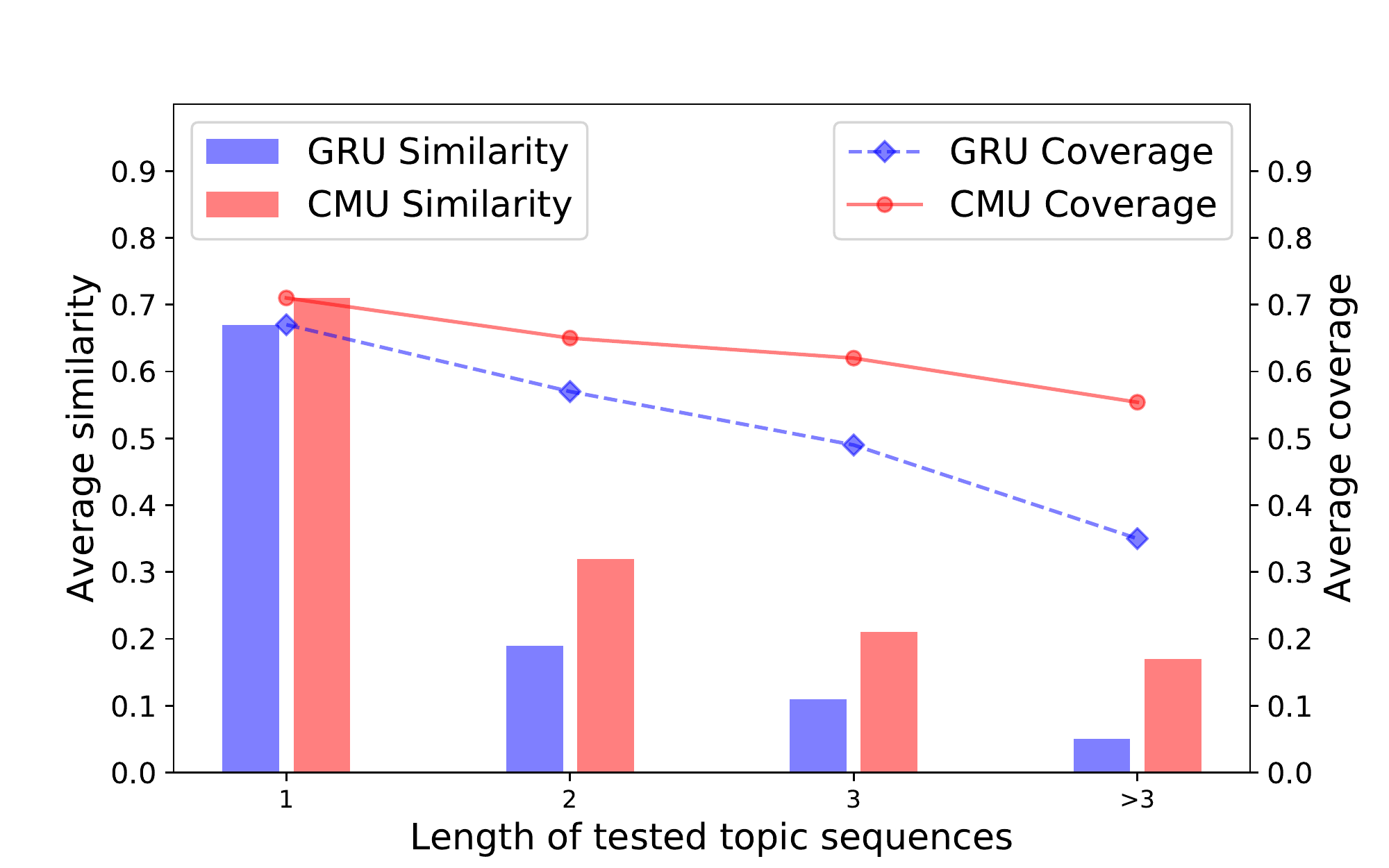}
  \caption{Performance comparison of CMU and GRU in topic reordering.
  CMU achieves 1.41x improvement in similarity and 1.20x in coverage over GRU.}
  \label{fig:seq_eval}
  \vspace{-0.2cm}
\end{figure}
\subsection{Ablation Studies on Sentence Generation}
\label{sec:study_sg}
In this subsection, we compare the performance of CSG with the state-of-the-arts on data-to-text generation benchmarks.

\fakeparagraph{Baselines}
We compare our model with three state-of-the-arts where first two do not have topic planning while PHVM does:
\begin{itemize}
  \item Seq2Seq+Attention \cite{sutskever2014sequence}.
  We use Seq2Seq model with attention as the baseline.
  An attention is added for Seq2Seq to focus on the given topic.
  \item CVAE \cite{sohn2015learning}.
  It is the basic structure of CSG.
  The topic serves as the context $c$ of CVAE.
  \item PHVM \cite{shao2019long}.
  It is also based on CVAE but utilizes the average pooling of ordered topics to predict the next one.
\end{itemize}

\fakeparagraph{Metrics}
We use the following metrics to evaluate the performance of our model and selected methods.
\begin{itemize}
  \item Similarity/Coverage:
  We adopt the metric used in \cite{li2018generating, shao2019long} to measure the consistency between (a) predicted topic sequence and the ground truth and (b) generated sentence and the target sentence.
  For similarity, we use sequence coincidence for topic sequences and cosine of vectors for sentences.
  For coverage, we use the ratio of overlapped words.
  We use k=8 as the top-k value from experiences.
  \item BLEU score:
  it is a widely used metric for measuring the overlap between the ground truth and generated sentences \cite{shao2019long, li2018generating}.
  \item Coherence and Relevance:
  these two metrics measure the quality of generated sentences, which are evaluated by human experts \cite{feng2018topic}.
\end{itemize}

\fakeparagraph{Setups}
We use the ads dataset and insurance dataset to evaluate the performance of CSG and other baselines in sentence generation.
Similar to the evaluation of topic recommendation, we use 80\% of data for training and the remaining 20\% for testing.


\fakeparagraph{Results}
Since the baselines are incapable of such fine-grained topic reordering, we prepare topic sequences for the baselines and evaluate their performance on the same topic sequences.
The results are shown in \tabref{table:sent_gen}.
Compared with the states-of-the-arts, CSG achieves 0.71\%-5.6\% improvement in BLEU and similarity, and 0.72\%-1.6\% improvement in coherence and relevance.
Note that CSG notably outperforms basic CVAE in all the metrics.
This indicates that taking supplementary information, \eg previous sentences, in context improves overall quality of the generated sentences.
In addition, concatenating the embedded vectors of topic and previous sentence enables CSG to generate more topic-relevant sentence while ensuring global coherence.

\begin{table*}[t]
\tiny
\caption{Performance comparison on sentence generation.
Coherence and Relevance are evaluated by human experts.
BLEU scores are automatically calculated.
Relative improvement with the best candidate is marked bold in the last row.}
\centering
\resizebox{\textwidth}{!}{%
\begin{tabular}{l|l|l|l|l|l|l|l|l|l|l}
\hline
\multicolumn{1}{c|}{\multirow{2}{*}{Model}}              & \multicolumn{5}{c|}{Ads dataset}                                                                                                                                                                                                                                                                      & \multicolumn{5}{c}{Insurance dataset}                                                                                                                                                                                                                                                                   \\ \cline{2-11}
\multicolumn{1}{c|}{}                                    & \textbf{BLEU-1}                                            & \textbf{BLEU-4}                                           & \textbf{Similarity}                                             & \textit{\textbf{Coherence}}                             & \textit{\textbf{Relevance}}                             & \textbf{BLEU-1}                                           & \textbf{BLEU-4}                                            & \textbf{Similarity}                                              & \textit{\textbf{Coherence}}                              & \textit{\textbf{Relevance}}                              \\ \hline
Seq2Seq+Att                                               & 0.2078                                                    & 0.0271                                                    & 0.3152                                                   & 3.51                                                    & 3.76                                                    & 0.2311                                                     & 0.0421                                                     & 0.4235                                                    & 3.94                                                     & 4.16                                                     \\ \hline
CVAE                                                     & 0.1970                                                     & 0.0246                                                   & 0.2714                                                   & 3.42                                                    & 3.14                                                    & 0.2194                                                    & 0.0274                                                     & 0.4204                                                    & 3.25                                                     & 3.94                                                     \\ \hline
PHVM                                                     & 0.2104                                                    & 0.0279                                                    & 0.3421                                                    & 3.78                                                    & 3.91                                                    & 0.2532                                                    & 0.0457                                                     & 0.4671                                                    & 4.13                                                     & 4.32                                                     \\ \hline
\begin{tabular}[c]{@{}l@{}}SG(CSG)\\ RelImp\end{tabular} & \begin{tabular}[c]{@{}l@{}}0.2119\\ (\textbf{+0.71\%})\end{tabular} & \begin{tabular}[c]{@{}l@{}}0.0282\\ (\textbf{+1.1\%})\end{tabular} & \begin{tabular}[c]{@{}l@{}}0.3530\\ (\textbf{+2.9\%})\end{tabular} & \begin{tabular}[c]{@{}l@{}}3.84\\ (\textbf{+1.6\%})\end{tabular} & \begin{tabular}[c]{@{}l@{}}3.97\\ (\textbf{+1.5\%})\end{tabular} & \begin{tabular}[c]{@{}l@{}}0.2564\\ (\textbf{+1.2\%})\end{tabular} & \begin{tabular}[c]{@{}l@{}}0.0461\\ (\textbf{+0.87\%})\end{tabular} & \begin{tabular}[c]{@{}l@{}}0.4935\\ (\textbf{+5.6\%})\end{tabular} & \begin{tabular}[c]{@{}l@{}}4.16\\ (\textbf{+0.72\%})\end{tabular} & \begin{tabular}[c]{@{}l@{}}4.36\\ (\textbf{+0.92\%})\end{tabular} \\ \hline
\end{tabular}
}
\label{table:sent_gen}
\end{table*}


\begin{figure}[t]
  \centering
  \includegraphics[width=0.43\textwidth, height=2.7in]{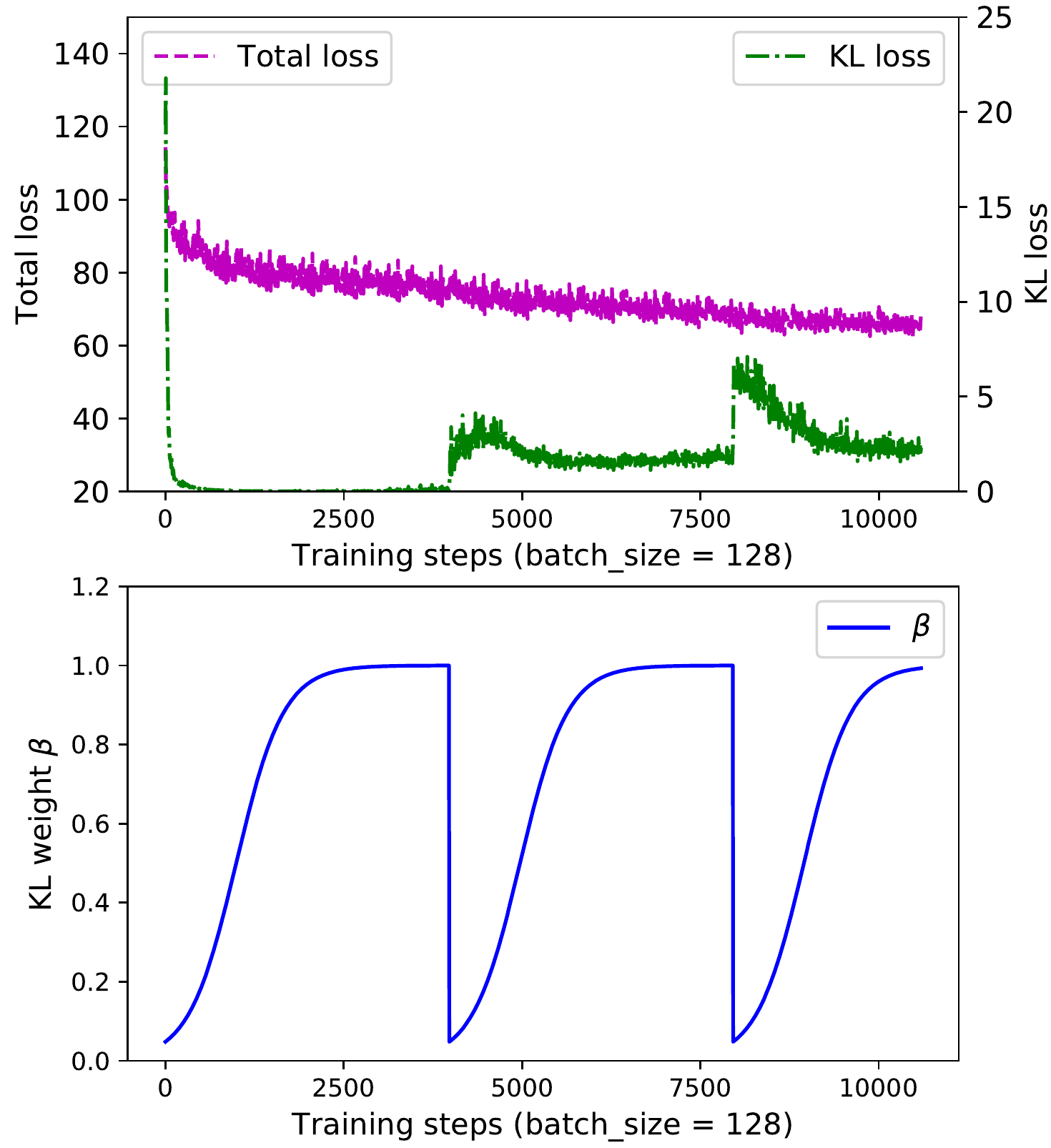}
  \caption{An example of our intercepted cyclical training scheme on KL-vanishing problem. Training set is insurance dataset. The upper figure refers to the loss in the training, while the bottom one refers to the controlled KL weight $\beta$.}
  \label{fig:kl}
  \vspace{-0.2cm}
\end{figure}

\subsection{Additional Ablation Studies of CSG}

In subsection, we present a more detailed study of CSG on sentence generation, \ie its ability in generating attractive sentences.
First, we evaluate if CSG can reorder topics according users' preferences (see \secref{sec:reorder}).
Then we check the performance of the dynamic control scheme in mitigating KL-vanishing problem (see \secref{sec:kl_vanish}).
This is important for the diversity of generated sentences and avoiding getting users bored.

\subsubsection{Topic Reordering}
\label{sec:reorder}
As mentioned before, we first evaluate the performance of CSG in topic reordering, \ie the performance of context management unit (CMU) in CSG.
We split the test samples into different groups based on the length of topic sequences.
As shown in \figref{fig:seq_eval}, CMU outperforms GRU in both the similarity and the coverage of generated topic sequence on ground truth.
This is because the diverse user preferences place different attention weights on topics even in the same topic set, which results in various topic sequences.
It can be observed from the gap between coverage and similarity, since only the similarity metric considers the order in calculation.
To address this issue, CMU takes user's feature embedding in reordering topics and thus achieves better performance than GRU, which solely learns grammatical transition probability from topic sequences.

\subsubsection{Mitigating KL-Vanishing Problem}
\label{sec:kl_vanish}
Next we demonstrate the results of dynamic controlling scheme of $\beta$ in mitigating KL-vanishing problem.
Since our sentence generator is built upon CVAE, the KL-vanishing problem should be addressed in the training process.
KL-vanishing problem is due to the powerful learning ability of autoencoder, \ie it can directly learn $w^{\bm{S}_{i}}_j$ conditioned on $w^{\bm{S}_{i}}_{<j}$, making $\bm{z}$ independent of $(\bm{enc(S)}, c)$.
At this time, $\mathcal{L}_{KL}$ vanishes and the model cannot generate context-aware sentences \cite{sohn2015learning}.
Enlightened by \cite{fu2019cyclical}, we propose an intercepted cyclical $\beta$ controlling scheme for mitigating KL-vanishing problem, \ie \equref{eq:beta_control}.
The cycle makes $z$ able to capture more structured information and facilitates CSG in learning sentence diversity conditioned on $(\bm{enc(S)}, c)$.
The interception reduces the initial KL weight $\beta$, which can better initialize $z$ for further training.
We use insurance dataset as an example and the parameters are shown in \tabref{table:sg_parameter}.
As shown in \figref{fig:kl}, the cyclically increased $\beta$ steadily increases KL-loss for each cycle.
This enables CSG to learn more about the diversity of sentences for generation.

\section{Online Evaluation}
\label{sec:online}
This section reports the online evaluation of \sysname on a large online insurance platform.

\fakeparagraph{Setups}
The online experiments are setup as follows: we first conduct validation test on a small group of junior agents and senior agents.
The test examines the impacts of \sysname on those two groups separately.
Then we deploy \sysname to all agents as global test to validate the effectiveness of \sysname in increasing total premium.
Following the global test, we conduct a case study on reached users in global test to further understand how \sysname stimulates their interests of insurance purchase.

\begin{figure}[t]
  \centering
  \includegraphics[width=0.43\textwidth, height = 1.5 in]{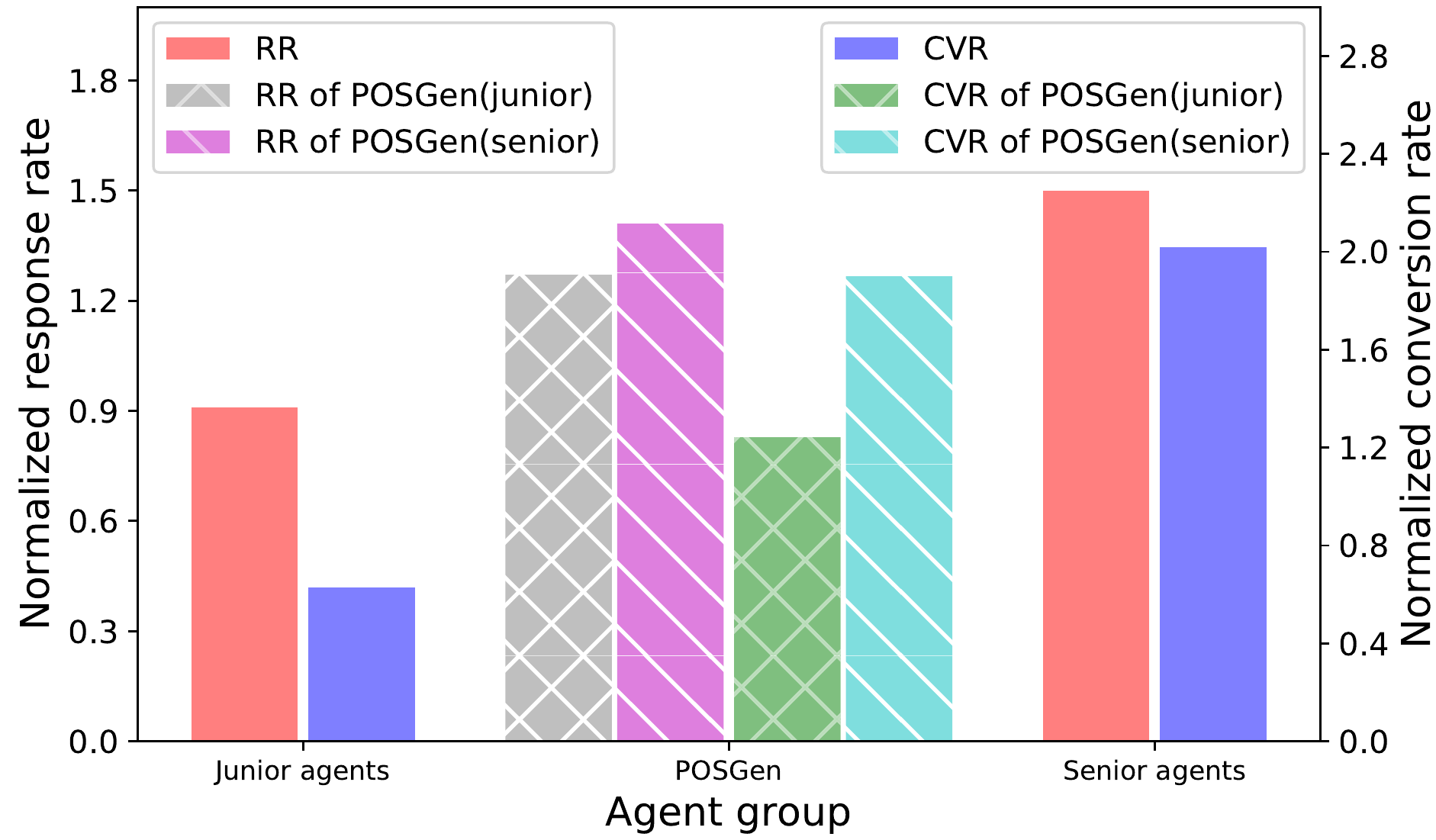}
  \caption{Validation test results.
  \sysname improves the performance of junior agents by 1.4x in RR and 1.73x in CVR and achieves competitive results to senior agents.}
  \label{fig:online_eval}
  \vspace{-0.2cm}
\end{figure}

\begin{figure*}[t]
  \centering
  \includegraphics[width=0.96\textwidth,height=1.5in]{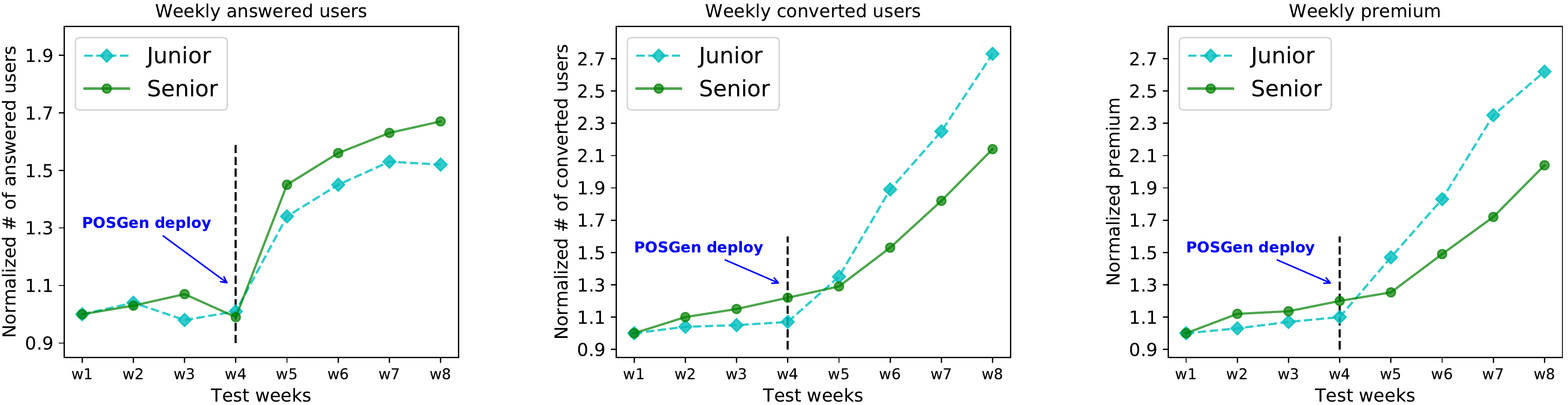}
  \caption{Global test results.
  For better comparison, we evaluate junior and senior agents separately after deployment.
  }
  \label{fig:large_deployment}
\end{figure*}
\subsection{Validation Test}
\fakeparagraph{Metrics}
In the validation test, we randomly replace a group of human agents by \sysname to reach out for customers, and those who respond to the opening sentences generated by \sysname are handed over to human agents.
We use response rate and conversion rate as metrics for the validation test.


\fakeparagraph{Results}
\figref{fig:online_eval} shows the results of validation test.
For better illustration, the test group is split into junior and senior.
We observe notably improvement by \sysname for the junior group.
Specifically, \sysname improves junior's RR by 1.4x and CVR by 1.73x.
Looking into the response data, we find that the greater improvement in CVR is due to the higher ratio of affirmative responses brought by \sysname.
Note that \sysname also achieves similar performance as the senior group.
Compared with senior agents, the gap is around 10\%.
These results demonstrate that \sysname can greatly improve the efficiency of agents in approaching potential customers while guaranteeing the conversion rate.
It implies that \sysname may also increase the overall premium brought by the agents, which leads to the global test below.

\subsection{Global Test}
\fakeparagraph{Metrics}
The global test includes all agents and we evaluate the improvement in total answered users, converted users and premium.
The test lasts two months.

\fakeparagraph{Overall performance}
\figref{fig:large_deployment} show the results.
Among three evaluated metrics, the number of weekly answered users immediately improved after deployment.
The average improvement from w5 to w8 is 1.52x, where 1.46x is for junior agents and 1.58x for senior.
This is because \sysname is effective in stimulating user responses and can reach more potential customers than human agents.
For weekly converted users and premium, the improvement is lagged for around one week.
This is because a large portion of users need time to make buying decisions.
The gain on weekly converted users and premium is also significant.
For converted users, the growth is greatly accelerated, \ie the improvement of average weekly growth is 11.1x for junior and 2.1x for senior.
Based on the growth before \sysname is deployed, the estimated average improvement in converted users is 1.91x.
The improvement on weekly premium shows the same pattern, \ie 2.33x improvement in average.
The results demonstrate the effectiveness of \sysname in promoting online insurance sales.

\begin{figure}[t]
  \centering
  \includegraphics[width=0.43\textwidth, height = 1.7 in]{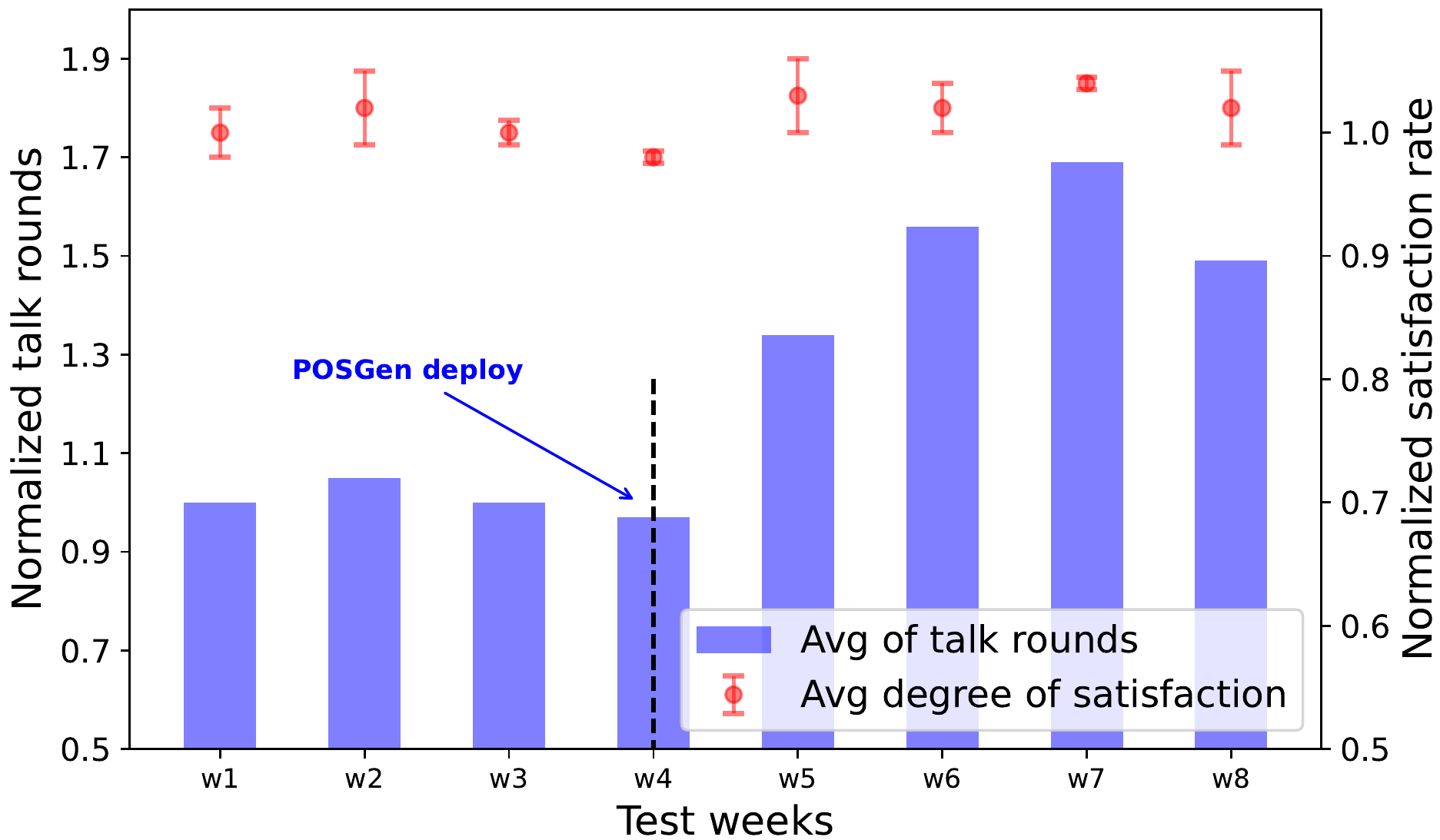}
  \caption{Quality of users' experiences (QoE) analysis in global test. After deployment, the average talk rounds increased 51\%. The average degree of satisfaction is slightly increased by 2\%. }
  \label{fig:satisfaction}
  \vspace{-0.2cm}
\end{figure}

\fakeparagraph{Performance on QoE}
In addition, we also review the impact of \sysname on the quality of users' experiences (QoE).
Consist with our commercial practice, we utilize average talk rounds and the degree of satisfaction as two quantifying metrics.
The average talk rounds are automatically derived from our data warehouse.
For the degree of satisfaction, it is calculated based on users' feedback and human online check.
As shown in \figref{fig:satisfaction}, the average talk rounds are increased by 51\% after the deployment.
The results show that finding users' most concerned topics is more effective in raising their interests to continue the talk than solely focusing on selling products.
In addition, we also notice that the degree of satisfaction is also slightly increased by 2\%.
In summary, the global test proves \sysname's effectiveness in increasing the insurance sales of agents without decreasing the quality of users' experiences (QoE).

\subsection{Case Study on Users in Global Test}

\begin{table}[htbp]
\centering
\caption{Classifying metrics on users' response.}
\resizebox{\columnwidth}{!}{%
\begin{tabular}{|l|l|l|l|}
\hline
\multicolumn{1}{|c|}{\textbf{Group}} & \multicolumn{1}{c|}{\textbf{Example}}                    & \multicolumn{1}{c|}{\textbf{AutoEval}}                         & \multicolumn{1}{c|}{\textbf{HumanEval}} \\ \hline
Affirmative                          & 1; OK                                                & Keyword match                                                  & Yes                                          \\ \hline
Questionable                          & \begin{tabular}[c]{@{}l@{}}Could you tell me \\ the detailed terms? \end{tabular} & \begin{tabular}[c]{@{}l@{}}Keyword match\\ +model check\end{tabular} & Yes                                          \\ \hline
Negative                             & No need; No thanks                                                   & Keyword match                                                  & Yes                                          \\ \hline
No response                          &                                                          & T+3 check                                                       & No                                           \\ \hline
\end{tabular}
}
\label{table:metrics}
\end{table}

Finally, we conduct a case study on the reached users in the global test.
The case study provides details to understand how \sysname improves the final insurance sales.

\fakeparagraph{Metrics}
We randomly select 10000 users touched by \sysname (after week 4) and agents (before week 4) for evaluation.
The evaluation consists of automatic evaluation, \eg pattern mining, and human evaluation.
The responses are classified into four groups: affirmative, questionable, negative and no response.
Affirmative and negative groups refer to whether a user wants to continue the conversation.
Questionable group refers to the responses containing specific questions.
Note that we mark a user as \emph{no response} if he/she did not respond in three days.
We detail the evaluation metrics in \tabref{table:metrics}.

\fakeparagraph{Results}
\tabref{table:response_study} lists the results.
The table contains two metrics, \ie normalized proportion improvement and normalized CVR improvement.
Compared with the responded users touched by agents, the proportions of affirmative group and questionable group ares greatly increased, \ie 370\% for affirmative group and 210\% for questionable group.
The results validate the effectiveness of \sysname in raising users' interests.
In addition, we also see that the responded users in affirmative group and questionable group of \sysname have similar Normalized CVR to those of agents.
This observation implies that \sysname is able to find more potential users with purchase intention, which are missed by generic opening sentences.
The decrease of negative and ``no response'' users also validate the results.


\begin{table}[htbp]
\centering
\caption{Case study on users' response of POSGen and agents. Normalized CVR improvement is also listed for each group.}
\resizebox{\columnwidth}{!}{%
\begin{tabular}{|l|l|l|}
\hline
\multicolumn{1}{|c|}{\textbf{Group}} & \multicolumn{1}{c|}{\textbf{\begin{tabular}[c]{@{}c@{}}Normalized Proportion Impr\\ (POSGen vs Agents)\end{tabular}}} & \multicolumn{1}{c|}{\textbf{\begin{tabular}[c]{@{}c@{}}Normalized CVR Impr \\ (POSGen vs Agents)\end{tabular}}} \\ \hline
Affirmative                          & 370\%                                                                                                               & 5\%                                               \\ \hline
Questionable                         & 210\%                                                                                                               & 21\%                                              \\ \hline
Negative                             & -12\%                                                                                                               & -0.4\%                                            \\ \hline
No response                          & -11\%                                                                                                               & 0.1\%                                             \\ \hline
\end{tabular}
}
\label{table:response_study}
\end{table}

\section{Conclusion}
\label{sec:conclusion}
In this paper, we design POSGen, a personalized opening sentence generation scheme for promoting online insurance sales.
POSGen consists of two components: selective attentive transfer learning (SATL) model and context-aware sentence generator (CSG).
SATL learns users' favorite topics with limited historical conversation samples by transferring knowledge from auxiliary samples of users' behaviors.
The learned topics and user embeddings are then passed to CSG to generate opening sentences with user-specific topic ordering.
Extensive offline and online experiments show that POSGen is effective in generating personalized opening sentences and promoting online insurance sales. 
\clearpage

\bibliographystyle{IEEEtran}
\bibliography{reference}

\end{document}